\documentclass[runningheads]{llncs}
\pdfoutput=1 
 
\usepackage{eccv}

\usepackage{eccvabbrv}

\usepackage{graphicx}
\usepackage{booktabs}

\usepackage[accsupp]{axessibility}  

\usepackage{hyperref}

\usepackage{orcidlink}

\usepackage{tcolorbox}
\usepackage{multirow}
\usepackage[ruled]{algorithm2e}
\usepackage[normalem]{ulem}
\useunder{\uline}{\ul}{}
\usepackage{listings}
\usepackage{wrapfig}
\usepackage{amssymb} 
\SetSideCommentRight

\SetCommentSty{mycommfont}
\usepackage{xcolor}
\definecolor{verylightgray}{gray}{0.9} 
\usepackage{indentfirst}
\usepackage{markdown}

\begin{document}

\title{DEAL: Disentangle and Localize Concept-level Explanations for VLMs} 


\author{Tang Li \orcidlink{0000-0002-3134-4151} \and
Mengmeng Ma \orcidlink{0000-0002-2804-2718} \and
Xi Peng \orcidlink{0000-0002-7772-001X}}

\authorrunning{T. Li et al.}

\institute{Deep-REAL Lab: \url{https://deep-real.github.io/} \\
Department of Computer \& Information Science, University of Delaware \\
\email{\{tangli,mengma,xipeng\}@udel.edu}}

\maketitle

\begin{abstract}
Large pre-trained Vision-Language Models (VLMs) have become ubiquitous foundational components of other models and downstream tasks. Although powerful, our empirical results reveal that such models might not be able to identify fine-grained concepts. Specifically, the explanations of VLMs with respect to fine-grained concepts are entangled and mislocalized. To address this issue, we propose to DisEntAngle and Localize (DEAL) the concept-level explanations for VLMs without human annotations. The key idea is encouraging the concept-level explanations to be distinct while maintaining consistency with category-level explanations. We conduct extensive experiments and ablation studies on a wide range of benchmark datasets and vision-language models. Our empirical results demonstrate that the proposed method significantly improves the concept-level explanations of the model in terms of disentanglability and localizability. Surprisingly, the improved explainability alleviates the model's reliance on spurious correlations, which further benefits the prediction accuracy.
\end{abstract}
\section{Introduction}
\label{sec:intro}
\footnotetext[1]{Our source code and pretrained weights: \href{https://github.com/tangli-udel/DEAL}{https://github.com/tangli-udel/DEAL}.
}

\begin{figure}[t]
  \centering
   \includegraphics[width=1.0\linewidth]{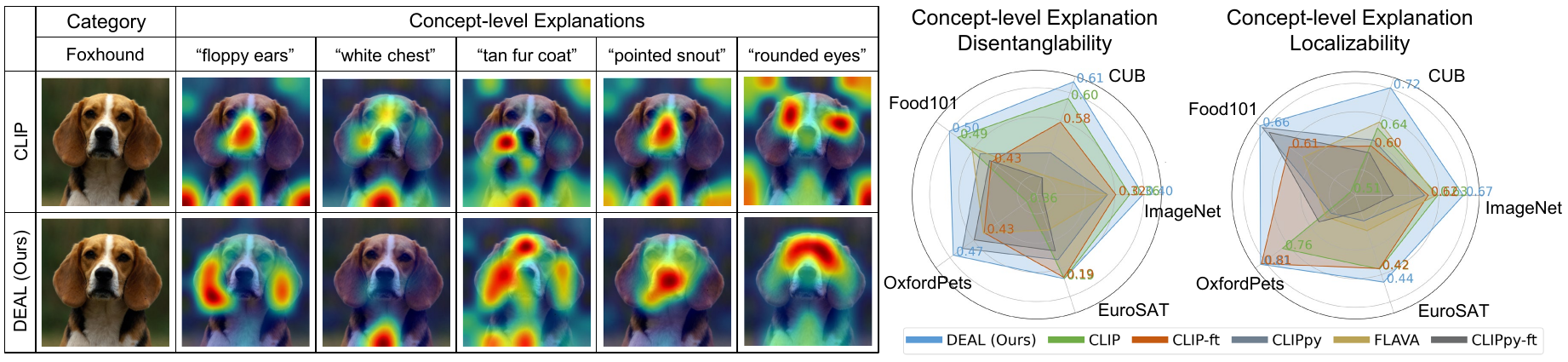}
   \caption{
   We visualize the CLIP~\cite{radford2021learning} model's explanations for fine-grained concepts.
   The comparison is conducted using Chefer et al.~\cite{Chefer_2021_ICCV} explanation method on the ViT-B/32 vision backbone.
   ({\it Left}) CLIP's explanations {\it w.r.t.} visually distinct concepts might highlight the same region, or even mislocalize concepts to spurious factors, {\it e.g.}, background.
   In contrast, the concept-level explanations of our model are well-disentangled and correctly localized.
   ({\it Right}) Our method significantly improves the model's concept-level explanations in terms of disentanglability and localizability compared to state-of-the-art VLMs on different benchmark datasets.
   This figure is best viewed in color.
   }
   \label{fig:title}
\end{figure}

\noindent Since the introduction of large pre-trained Vision-Language Models (VLMs), {\it e.g.}, CLIP~\cite{radford2021learning} and FLAVA~\cite{singh2022flava}, they become ubiquitous ``foundations'' for other models~\cite{bommasani2021opportunities, liu2024visual} and downstream tasks~\cite{shen2021much, li2022blip}.
Although these models exhibit significant capabilities, our empirical results reveal that they might not be able to identify fine-grained concepts.
As shown in Fig.~\ref{fig:title}, VLMs often associate related concepts to the main object without further distinctions.
This would yield severe consequences in safety-critical downstream tasks.
For example, in autonomous vehicles, a model that entangles shapes, colors, or contexts with specific road signs, might misinterpret or fail to recognize temporary or region-specific signs.
This could lead to unsafe driving decisions, endangering passengers and other road users.
A question naturally arises: {\it Can we develop VLMs that disentangle and localize fine-grained linguistic concepts on images}?

This task is non-trivial due to: (i) learning objective mismatch and (ii) supervision gap.
On the one hand, the learning objective of VLMs aims to align the image with the entire description, rather than distinguishing individual concepts within the description~\cite{he2020momentum, radford2021learning, thrush2022winoground}.
This entangles the model's understanding of different concepts~\cite{yamada2022lemons, yuksekgonul2022and, ma2023crepe}.
On the other hand, existing works for finer-grained alignment of VLMs rely on human annotations, {\it e.g.}, bounding boxes, to localize different objects~\cite{zeng2022multi, li2022grounded}.
However, their supervision is limited to the object-level ({\it i.e.}, category-level) granularity.
Obtaining human annotations at concept-level is prohibitively expensive, or even impossible~\cite{wang2020self, li2023data}.
Therefore, how to train VLMs that can localize fine-grained concepts remains an open question.

In this paper, we propose a straightforward but effective method to {\bf D}is{\bf E}nt-{\bf A}ngle and {\bf L}ocalize (DEAL) the concept-level explanations for solving this challenging task.
The key idea is to harness the discrepancy and consistency between explanations at the concept and category levels for self-supervised learning.
This will enable us to generate supervisory signals for concept learning without manual annotations.
Our method works as a plug-in module, seamlessly integrating with the existing model architectures without structural changes.
After training, our method provides explanations for the predictions using human-understandable concepts based on reliable visual evidence.

We evaluate the proposed method on a wide range of benchmark datasets and different vision backbones.
The extensive comparisons and ablation studies prove the effectiveness of the proposed method.
Quantitatively, we propose two metrics, concept-level explanation disentanglability and localizability, to evaluate the explanation quality without human annotations.
The empirical results demonstrate the superiority of our concept-level explanations on both metrics.
Qualitatively, in contrast to existing VLMs that exhibit an entangled understanding of concepts, our model can visually disentangle and correctly localize fine-grained concepts.
The results indicate that our model understands the semantic meanings of concepts even though their appearance drastically changed across categories.
Furthermore, our model exhibits superior performance in localizing object parts, as validated by ground truth segmentation masks.

Surprisingly, the proposed method also significantly improves VLM's prediction accuracy.
This is different from most interpretability methods that come with a compromised benchmark performance~\cite{koh2020concept, chen2019looks}.
We argue that the disentangled concepts would alleviate the model's reliance on spurious correlations~\cite{qiao2020learning, ma2024beyond}, thereby enabling better generalization capability to unseen data.

In summary, our main contributions:
\begin{itemize}
  \item Our empirical results reveal that the existing VLMs cannot distinguish between fine-grained concepts.
  Their explanations with respect to fine-grained concepts are entangled and mislocalized.

  \item We propose a straightforward but effective method, called DEAL, to disentangle and localize the concept-level explanations for VLMs without human annotations. 
  Our method works as a plug-in module without changing the model architecture.

  \item Empirical results in a wide range of benchmark datasets and VLMs demonstrate that our method significantly improves the concept-level explanation in terms of disentanglability and localizability by 8.8\% and 10.9\% on average.
  Furthermore, the enhanced explainability also improves the model's prediction accuracy by 1.6\% on average.
\end{itemize}

\section{Related Works}
\label{sec:related_works}
{\bf eXplainable Machine Learning (XML).}
The emerging field of XML aims to bring transparency to today’s powerful but opaque deep learning models.
XML usually consists of two categories.
(1) Intrinsic methods, whose explanations are inherent to the model's design and training, provide explanations along with output, such as joint training~\cite{hind2019ted,ma2021smil} and prototyping~\cite{chen2019looks} methods.
However, such interpretable methods usually sacrifice their prediction performance~\cite{rudin2019stop, chen2019looks}.
(2) Post hoc methods, which give insight into the learned associations of a model that are not readily interpretable by design.
Such methods usually leverage backpropagation or local approximation to offer saliency maps as explanations, {\it e.g.}, Vanilla Gradient~\cite{simonyan2013deep}, Grad-CAM~\cite{selvaraju2017grad}, LIME~\cite{ribeiro2016should}, and SHAP~\cite{lundberg2017unified}.
However, such saliency-based explanations are inherently ambiguous and require ML expertise to understand~\cite{rudin2019stop, li2021deep}.
For example, given a saliency map that highlights the correct object on the image, it remains unclear whether the model's prediction stems from the texture or color.
In contrast, the proposed method offers human-understandable explanations while maintaining high prediction performance.

{\bf Concept-based explanations.}
Instead of delving directly into raw features, there have been existing attempts to correlate the internal representations of machine learning models with high-level, semantically meaningful concepts familiar to humans.
They either provide quantification of the influence of predefined concepts like ``stripes'' on the model's predictions, {\it e.g.}, TCAV~\cite{kim2018interpretability}, Concept Bottleneck~\cite{koh2020concept}, or manually translate the abstract feature maps learned by the model into human-understandable terms, {\it e.g.}, ProtoPNet~\cite{chen2019looks}, CRP~\cite{achtibat2022towards}.
However, their explanations might not be faithful to the model.
The former relies on subjective user-defined concepts that might not align with the model's understanding, and is unable to localize their presence~\cite{havasi2022addressing}.
The latter is limited by human bias, it is impractical to interpret the vast and complex feature spaces of deep models~\cite{meng2022interpretability}.

\begin{figure}[t]
  \centering
   \includegraphics[width=0.95\linewidth]{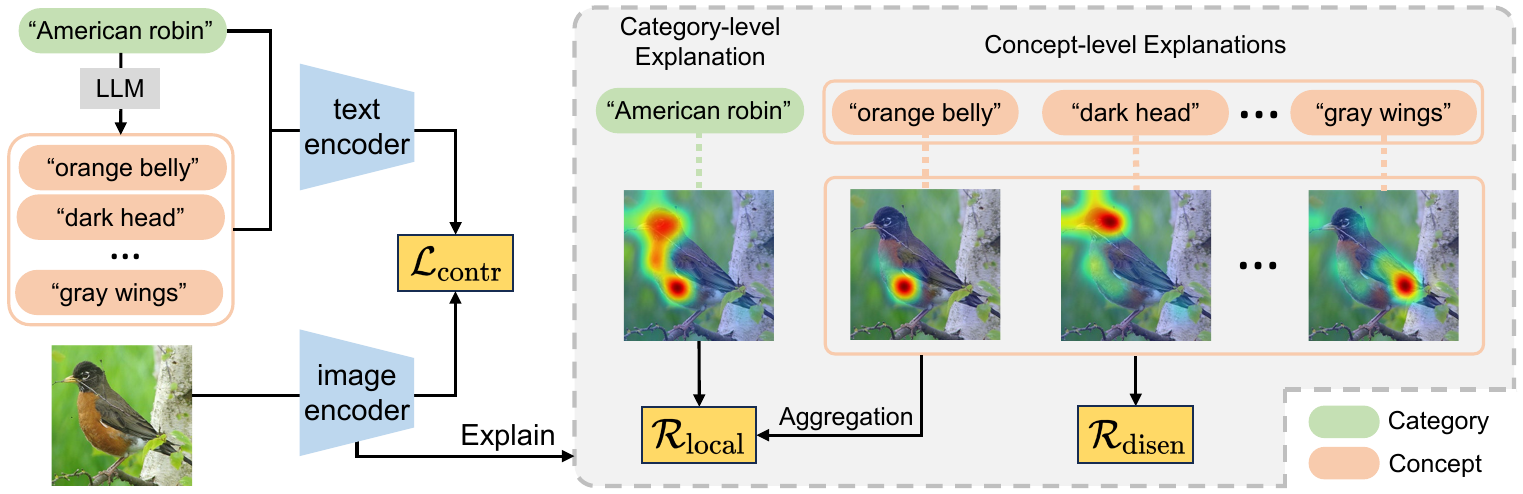}
   \caption{
   {\bf Overview of the proposed method to Disentangle and Localize (DEAL) concept-level explanations for VLMs.}
   Our method consists of the concept-level explanation disentanglement and localization constraints for training Vision-Language Models (VLMs).
   First, we query the Large Language Model (LLM), {\it e.g.}, GPT-3.5~\cite{peng2023gpt35turbo}, with the category name to obtain discriminative visual concepts for distinguishing the category.
   Then, we feed the category and concepts as a sentence into the text encoder, and the image into the image encoder to calculate the standard contrastive learning loss ($\mathcal{L}_\mathrm{contr}$).
   Next, we calculate the explanations {\it w.r.t.} the category name and each of the concepts within the category.
   We constrain the disentanglement between the explanations of all concepts within the category ($\mathcal{R}_\mathrm{disen}$).
   Finally, we constrain localization by the consistency between the category-level explanation and the aggregation of concept-level explanations ($\mathcal{R}_\mathrm{local}$).
   This figure is best viewed in color.
   }
   \label{fig:overview}
\end{figure}

{\bf Vision-Language Models (VLMs).}
Vision-Language Models have surged in popularity as they aim to bridge the domains of computer vision and natural language processing.
These models are designed to understand descriptions for visual content.
OpenAI's CLIP~\cite{radford2021learning} and Google's ViLBERT~\cite{lu2019vilbert} have showcased the potential of combining vision and language pre-training on large-scale datasets, resulting in state-of-the-art performance across a variety of benchmarks.
Decomposing categories into visual descriptors can further improve the prediction accuracy of VLMs~\cite{menon2022visual, qin2022medical, yang2023language}.
However, recent works reveal that they might lack compositional reasoning to understand the objects and their relationships~\cite{parcalabescu2021seeing, yuksekgonul2022and, ma2023crepe}.
Our empirical results further prove that VLMs cannot distinguish fine-grained concepts, their explanations with respect to these concepts are entangled and mislocalized.
Although there are recent attempts on fine-grained alignment, {\it e.g.}, X-VLM~\cite{zeng2022multi} and PyramidCLIP~\cite{gao2022pyramidclip}, they heavily rely on expensive human annotations and are limited to category (or object) level granularity.
\section{Methods}
\label{sec:methods}
\noindent In this section, we first present how to prompt Large Language Models (LLMs) for discriminative concepts, then provide the problem formulation, and then introduce a novel method to disentangle and localize concept-level explanations.
Fig.~\ref{fig:overview} shows an overview of our method.

\subsection{Prompting LLMs for Discriminative Concepts}
\label{sec:prompt}
\noindent The recent breakthroughs in LLMs demonstrate their capability to provide commonsense knowledge in language~\cite{nori2023capabilities}.
Therefore, LLMs are implicit knowledge bases that can be easily queried with natural language, even for non-experts.
Inspired by recent works in language prompting~\cite{menon2022visual, yang2023language}, we can extract the desired concepts for each category from LLMs efficiently for large-scale datasets.

\begin{figure}[t]
  \centering
   \includegraphics[width=0.8\linewidth]{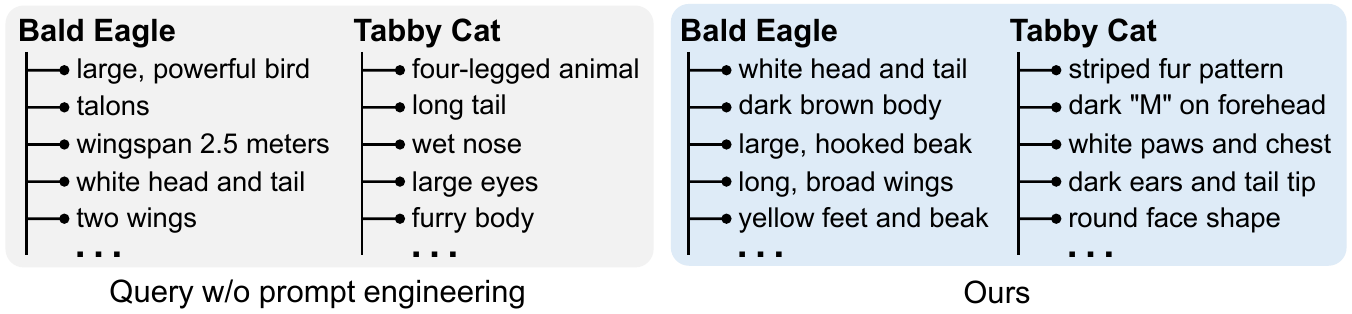}
   \caption{
   Directly query GPT-3.5~\cite{peng2023gpt35turbo} model would yield concepts that are not visually measurable.
   In contrast, our concepts are visually distinctable.
   The {\bf bold} words are the category names in {\it ImageNet}~\cite{deng2009imagenet}, and the list follows are the generated concepts.
   }
   \label{fig:example_concepts}
\end{figure}

However, as shown in Fig.~\ref{fig:example_concepts}, our empirical results show that the concept quality in terms of discriminative capacity is highly sensitive to the prompts.
To address this issue, we leverage the In-context learning~\cite{brown2020language} capability of LLMs to provide Chain-of-Thought (CoT)~\cite{wei2022chain} instructions.
Specifically, we provide the GPT-3.5~\cite{peng2023gpt35turbo} model with exemplary queries and responses, followed by the subsequent question:
\begin{lstlisting}[breakatwhitespace=true]
Q: What are the discriminative visual features with minimum overlap for identifying a [CATEGORY] in an image?
A: The discriminative visual features that identify a [CATEGORY] in an image are:
\end{lstlisting}
In contrast to direct query, our concepts align with the typical human rationale for predictions.
Fig.~\ref{fig:example_concepts} shows examples of our generated concepts for the corresponding category, which are more visually distinctable.

\subsection{Disentangling and Localizing Explanations}
\label{sec:deal}
{\bf Problem formulation.}
The contrastive learning objective of typical VLMs, {\it e.g.}, CLIP~\cite{radford2021learning}, is to learn a mapping function $f\in \mathcal{F}$ such that for each pair $(I, T) \sim P(\mathbf{I}, \mathbf{T})$, where $P(\mathbf{I}, \mathbf{T})$ is the distribution of image-text pairs, the function $f$  aligns the embeddings of $I$ and $T$ in a shared space if they are a correct match.
This process is typically guided by a contrastive loss function $\mathcal{L}_\mathrm{contr} (\cdot)$.
As we discussed in Secs~\ref{sec:intro} and~\ref{sec:related_works}, such learning objective is coarse-grained and ignores the distinction between fine-grained concept-level explanations.
To address this, we propose an effective approach to refine the learning process.

Let $g(\texttt{[TEXT]})$ be a post-hoc explanation method ({\it e.g.}, Grad-CAM~\cite{selvaraju2017grad}) that calculates the explanation heatmap for an image $I$ corresponding to the textual input \texttt{[TEXT]}, based on their similarity in the embedding space measured by $f$.
Considering all \texttt{[CONCEPT]} belong in to the same \texttt{[CATEGORY]}, we propose to {\it disentangle and localize (DEAL) concept-level explanations} by optimizing:
\begin{equation}
\begin{aligned}
\min_{f \in \mathcal{F}} \; & \mathrm {Risk}   (f) := \mathbb{E}_{(I,T) \sim P} \left[ \mathcal{L}_\mathrm{contr}(f(I, T)) \right] & & \lhd  {\textbf{\small Contrast}}\\
\text{s.t.} \; & \mathrm {Dist} (g({\texttt{\scriptsize [CONCEPT]}}), g({\texttt{\scriptsize [CONCEPT]}}'))  \ge \epsilon , & & \lhd  {\textbf{\small  Disentangle} }\\
&\mathrm {Dist} ( {\textstyle \sum} g({\texttt{\scriptsize [CONCEPT]}}), g({\texttt{\scriptsize [CATEGORY]}}))  \le \delta.  & & \lhd  {\textbf{\small Localize} }
\end{aligned}
\end{equation}
Intuitively, our goal is to ensure that the explanation of each concept within a category is distinct, while the collective explanations for all concepts are consistent with the category's overall explanation.
Next, we will explain the Contrast, Disentangle, and Localize in detail.

{\bf Contrastive learning objective.}
Following the standard contrastive learning methodology between image and text used in CLIP~\cite{radford2021learning}, we employ the InfoNCE loss~\cite{oord2018representation} as our backbone loss.
Denote $\left \{(I_i, T_i)  \right \}_{i=1}^{N}$ as a batch of $N$ paired images and texts, $(v_i, t_i) = f(I_i, T_i)$ as the vision and text embeddings, $\tau$ as a temperature scaling factor, then we define our contrastive learning loss as:
\begin{equation}
\mathcal{L}_\mathrm{contr}  = \frac{1}{N} \sum_{i=1}^{N} -\mathrm{log} \frac{\mathrm{exp}(v_i \cdot t_i / \tau ) }{ {\textstyle \sum_{m=1}^{N} \mathrm{exp}(v_i \cdot t_m / \tau )} }
\label{eq:contra}
\end{equation}
In contrast to the typical practice that simply uses category as the text description~\cite{wortsman2021robust, goyal2023finetune}, we augment them with the corresponding concepts obtained in Sec.~\ref{sec:prompt}.
For example, we can apply a template ``An image of \texttt{[CATEGORY]} with \texttt{[CONCEPT\_1]}, \texttt{[CONCEPT\_2]}, ..., and \texttt{[CONCEPT\_K]}''.

{\bf Disentangling concept-level explanations.}
As we previously discussed in Secs.~\ref{sec:intro} and~\ref{sec:related_works}, the key challenge for learning correct concept-level explanations is the absence of human annotations.
Let $\left \{C_k  \right \}_{k=1}^{K}$ be the set of $K$ concepts for the category of image $I$ as obtained in Sec.~\ref{sec:prompt}, then $g(C_{k})$ calculates the explanation heatmap for image $I$ with respect to concept $C_{k}$.
Denote $\mathrm {Dist} (\cdot, \cdot)$ as a distance metric, {\it e.g.}, $\ell_1$ distance, we propose to disentangle the concept-level explanations by imposing regularization on the model for each image using:
\begin{equation}
\mathcal{R}_\mathrm{disen} = - \sum_{k=1}^{K} \sum_{j=1, j\ne k}^{K} \mathrm {Dist} (g(C_{k}), g(C_{j}))  
\label{eq:disent}
\end{equation}
Intuitively, this regularization term constrains the model's explanations for each image {\it w.r.t.} different concepts to have minimum overlap.
Note that $g(\cdot)$ is not restrictive, including arbitrary gradient-based explanation implementation for VLMs, such as Chefer et al.~\cite{Chefer_2021_ICCV} and GradCAM~\cite{selvaraju2017grad}.
Different from existing methods that rely on category-level region annotations~\cite{zeng2022multi, li2022grounded}, our method provides finer-grained concept-level supervision without human annotations.

{\bf Localizing concept-level explanations.}
Our empirical results (Tab.~\ref{tab:ablation}) indicate that recklessly optimizing Eq.~(\ref{eq:disent}) could easily fall into local minimums.
For example, the explanations highlight well-separated parts of the image that are not related to the concepts.

``The meaning of the whole is a function of the meanings of its parts'' is held to be a key characteristic of human intelligence~\cite{cresswell1973logics, ma2023crepe}.
Following the same spirit, we propose to further regularize the concept-level explanations with corresponding category-level explanations.
Denote $g(\texttt{[CATEGORY]})$ as the explanation heatmap for image $I$ with respect to its category name, we localize the concept-level explanations by regularizing the model using:
\begin{equation}
\mathcal{R}_\mathrm{local} = \mathrm {Dist} (\frac{1}{K} \sum_{k=1}^{K}g(C_k), g(\texttt{[CATEGORY]})) 
\label{eq:consis}
\end{equation}
Intuitively, this regularization term constrains the aggregation of all concept-level explanations within the same category, to be consistent with the category-level explanation.
The proposed two constraints work in an adversarial-like manner.
This mechanism ensures distinct concept-level explanations are collectively coherent within their category, thereby forcing effective localization.

\subsection{Solving the Constrained Optimization}
\noindent Solving such a constrained optimization problem in Sec.~\ref{sec:deal} often leads to a non-convex problem, wherein methods like stochastic gradient descent (SGD) cannot guarantee constraint satisfaction.
To address this problem, we leverage Karush–Kuhn–Tucker conditions~\cite{boyd2004convex} and introduce Lagrange multipliers $\lambda$ and $\gamma$ to convert the constrained problem into its unconstrained counterpart.
Our overall learning objective is formulated as follows:
\begin{equation}
\min_{f \in \mathcal{F}}  \{\mathrm {Risk}(f) := \mathbb{E}_{(I,T) \sim P(\mathbf{I}, \mathbf{T})} \left[ \mathcal{L}_\mathrm{contr}(f(I, T)) \right] + \lambda \mathcal{R}_\mathrm{disen} +  \gamma  \mathcal{R}_\mathrm{local} \}
\label{eq:overall}
\end{equation}

\begin{algorithm}[t]
\caption{\footnotesize The proposed DEAL algorithm.}
\LinesNumbered
\label{alg:DEAL}
\KwIn{Image-Text distribution $P(\mathbf{I}, \mathbf{T})$;  step size $\eta$}
\KwOut{Learned VLM parameters $\theta$}
\For{all {\texttt{\footnotesize [CATEGORY]}} }{
    $\{{\texttt{\footnotesize[CONCEPT]}} \} \gets \mathrm{LLM} (\mathrm{prompt}({ \texttt{\footnotesize [CATEGORY]}})) $  \tcp*{Generate concepts}
}
\While{not converged}{
     Sample $(I,T) \sim P(\mathbf{I}, \mathbf{T})$  \tcp*{Sample Image-Text pairs}
     Calculate $\mathrm {Risk}(f)$ via Eq.~(\ref{eq:overall}) \tcp*{Calculate overall risk}
     Update $\theta$ via $\theta^{t+1} = \theta^{t} - \eta^{t}\nabla \mathrm{Risk}(f) $  \tcp*{Update model parameters}
}
\end{algorithm}
The entire training pipeline is summarized in Alg.~\ref{alg:DEAL}.
The proposed method has the following merits.
(1) In contrast to the pertaining objective of VLMs which is limited to coarse-grained alignment~\cite{he2020momentum, radford2021learning, thrush2022winoground}, the proposed learning objective can disentangle and localize fine-grained concepts. 
(2) Different from existing category-level alignment works that rely on human annotations~\cite{zeng2022multi, li2022grounded}, the proposed method leverages explanations to achieve an even finer, concept-level alignment in a self-supervised manner.
(3) The proposed method introduces no additional parameters, it works as a plug-in module for training VLMs via concept-level supervision without changing the model architecture.
\section{Experiments}
\label{sec:experiments}
\noindent To best validate the explanation quality and prediction performance, we conduct a series of experiments to compare the proposed DEAL method with other learning methods on different vision backbones.
The experimental results prove that our method achieves superior concept-level explanation disentanglability and localizability on a wide scope of benchmark datasets.
In contrast to most interpretability methods that come with a compromise on the benchmark performance, our method benefits the model's prediction.

\subsection{Datasets and Implementation Details}
{\bf Datasets.}
We conduct experiments on five typical image recognition datasets that have been used for evaluating the CLIP~\cite{radford2021learning} model.
(1) {\it ImageNet}~\cite{deng2009imagenet} dataset stands as one of the typical benchmarks in the field of computer vision.
The dataset spans 1,000 object classes and 1,281,167 training images from the WordNet hierarchy, ensuring a vast and varied semantic landscape.
(2) {\it CUB}~\cite{wah2011caltech}, short for Caltech-UCSD Birds-200-2011, is a highly specialized dataset designed to distinguish between subcategories of birds.
The dataset contains 11,788 images of 200 bird species.
(3) {\it Food-101}~\cite{bossard14} dataset is designed for the task of food recognition.
The dataset comprises 101,000 images split across 101 food categories.
(4) {\it Oxford-Pets}~\cite{parkhi12a} dataset includes a set of 7,349 images divided across 37 distinct pet categories, including a balanced collection of 12 dog breeds and 25 cat breeds.
(5) {\it EuroSAT}~\cite{helber2019eurosat} dataset is derived from Sentinel-2 satellite images, including a diverse range of European land use and land cover categories. 
The dataset contains 27,000 geo-referenced samples of 10 different classes.

{\bf Implementation details.}
For all datasets, we use the implementations of CLIP~\cite{radford2021learning} for different vision backbones.
Due to the computational cost of training CLIP from scratch, we focused on finetuning experiments using ViT-B/32 and ResNet-50 backbones.
We fine-tune the full model leveraging the proposed method, and compare the results with zero-shot and representative fine-tuning strategies of the CLIP model, including full model and vision-encoder-only fine-tuning.
For all the training, we use our augmented text of the category as the text description.
We split the training data of each dataset into 80\% and 20\% splits, and use the larger splits for training, and the smaller splits for validation and model selection.
We evaluate the model performance on the testing data provided by each dataset.
All models are trained using Adam~\cite{kingma2014adam} optimizer until convergence.
We crop the images of random size and aspect ratio, resizing to 224 $\times $ 224 pixels, random horizontal flips, random color jitter, grayscaling the image with 10\% probability, and normalization using the {\it ImageNet} channel statistics.
We use learning rate = 5e-7.
See Supplementary for more details.

\begin{table}[tb]
\centering
\caption
{
Comparison of explanation and prediction performance with state-of-the-art VLMs.
We conduct experiments on five datasets using the Vision-Transformer (ViT) encoders.
The models are tested on hold-out testing data.
Our results are on the average of three trials of experiments.
We highlight the {\bf best results} and the \underline{second best} results.
On average, our method exhibits superior concept-level explanation in terms of disentanglability and localizability.
The improved explainability also benefits the model's prediction accuracy.
$^\dagger$ Reported in~\cite{radford2021learning}, $^\ddagger$ reported in~\cite{li2021supervision}.
See Supplementary for the full table and the results using ResNet-50 backbone.
}
\resizebox{1.0\columnwidth}{!}{%
\begin{tabular}{clccccccc}
\toprule[1pt]
\multirow{2}{*}{Metrics}                                                                                          & \multicolumn{1}{c}{\multirow{2}{*}{Models}} & \multirow{2}{*}{\#Param.} & \multicolumn{6}{c}{Datasets}                                                                                                                 \\ \cmidrule{4-9} 
                                                                                                                  & \multicolumn{1}{c}{}                        &                           & ImageNet              & CUB            & Food101              & OxfordPets           & EuroSAT              & Avg.                           \\ \toprule[0.75pt]
\multirow{6}{*}{\begin{tabular}[c]{@{}c@{}}Concept-level Explanation\\ Disentanglability $\uparrow$\end{tabular}} & CLIP~\cite{radford2021learning}             & 151M                      & {\ul 0.361}           & {\ul 0.596}    & {\ul 0.487}          & 0.363                & \textbf{0.192}       & {\ul 0.400}                    \\
                                                                                                                  & FLAVA~\cite{singh2022flava}                 & 241M                      & 0.298                 & 0.541          & 0.463                & 0.429                & 0.115                & 0.369                          \\
                                                                                                                  & DeCLIP~\cite{li2021supervision}             & 186M                      & 0.032                 & 0.071          & 0.042                & 0.056                & 0.013                & 0.043                          \\
                                                                                                                  & PyramidCLIP~\cite{gao2022pyramidclip}       & 153M                      & 0.048                 & 0.116          & 0.080                & 0.085                & 0.026                & 0.071                          \\
                                                                                                                  & CLIPpy~\cite{ranasinghe2023perceptual}      & 196M                      & 0.300                 & 0.557          & 0.449                & {\ul 0.461}          & 0.162                & 0.386                          \\
                                                                                                                  & DEAL (Ours)                                 & 151M                      & \textbf{0.397}        & \textbf{0.608} & \textbf{0.501}       & \textbf{0.475}       & \textbf{0.192}       & \textbf{0.435}                 \\ \toprule[0.75pt]
\multirow{6}{*}{\begin{tabular}[c]{@{}c@{}}Concept-level Explanation\\ Localizability $\uparrow$\end{tabular}}    & CLIP~\cite{radford2021learning}             & 151M                      & 0.633                 & 0.638          & 0.511                & {\ul 0.762}          & {\ul 0.423}          & 0.593                          \\
                                                                                                                  & FLAVA~\cite{singh2022flava}                 & 241M                      & 0.630                 & 0.650    & 0.589                & 0.668                & 0.361                & 0.580                          \\
                                                                                                                  & DeCLIP~\cite{li2021supervision}             & 186M                      & 0.366                 & 0.367          & 0.318                & 0.369                & 0.295                & 0.343                          \\
                                                                                                                  & PyramidCLIP~\cite{gao2022pyramidclip}       & 153M                      & {\ul 0.662}           & {\ul 0.672}          & 0.644                & 0.700                & 0.302                & {\ul 0.596}                    \\
                                                                                                                  & CLIPpy~\cite{ranasinghe2023perceptual}      & 196M                      & 0.612                 & 0.614          & {\ul 0.656}          & 0.657                & 0.345                & 0.577                          \\
                                                                                                                  & DEAL (Ours)                                 & 151M                      & \textbf{0.673}        & \textbf{0.718} & \textbf{0.660}       & \textbf{0.809}       & \textbf{0.444}       & \textbf{0.661}                 \\ \toprule[0.75pt]
\multirow{6}{*}{Prediction Accuracy (\%)}                                                                         & CLIP~\cite{radford2021learning}             & 151M                      & $^\dagger$63.2        & {\ul 52.6}     & $^\dagger${\ul 84.4} & $^\dagger${\ul 87.0} & $^\dagger${\ul 41.1} & \multicolumn{1}{c}{{\ul 65.7}} \\
                                                                                                                  & FLAVA~\cite{singh2022flava}                 & 241M                      & 55.1                  & 49.4           & 79.7                 & 57.7                 & 28.2                 & 54.0                           \\
                                                                                                                  & DeCLIP~\cite{li2021supervision}             & 186M                      & $^\ddagger${\ul 66.2} & 35.9           & 57.0                 & 59.1                 & 27.3                 & 49.1                           \\
                                                                                                                  & PyramidCLIP~\cite{gao2022pyramidclip}       & 153M                      & 46.0                  & 43.8           & 49.3                 & 36.0                 & 20.0                 & 39.0                           \\
                                                                                                                  & CLIPpy~\cite{ranasinghe2023perceptual}      & 196M                      & 45.3                  & 18.9           & 53.8                 & 47.5                 & 18.9                 & 36.9                           \\
                                                                                                                  & DEAL (Ours)                                 & 151M                      & \textbf{70.8}         & \textbf{69.6}  & \textbf{86.9}        & \textbf{89.3}        & \textbf{77.4}        & \textbf{78.8}                  \\ \bottomrule[1pt]
\end{tabular}
}
\label{tab:experiments}
\end{table}

\begin{figure}[t]
  \centering
   \includegraphics[width=1.0\linewidth]{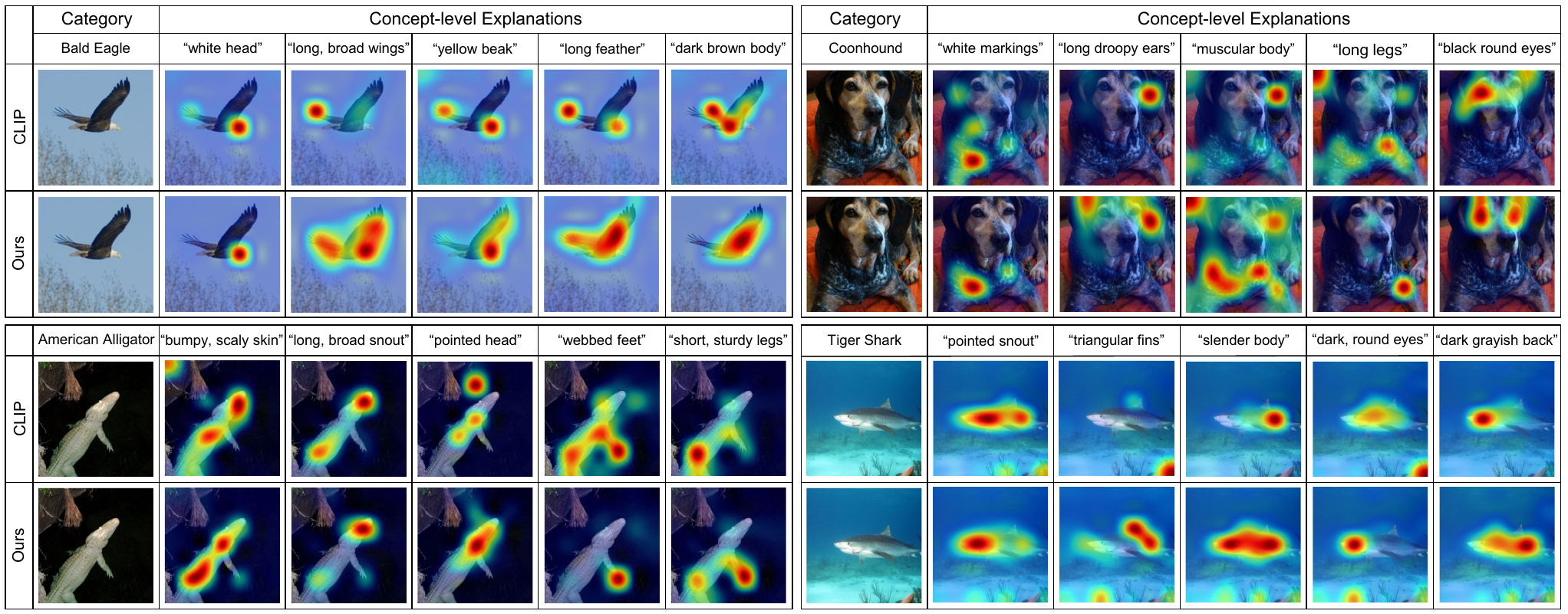}
   \caption{
   Disentanglability comparison using images from the {\it ImageNet}~\cite{deng2009imagenet} dataset.
   The CLIP~\cite{radford2021learning} model's explanations {\it w.r.t.} different concepts are entangled.
   They might highlight the same region for visually distinct concepts, or even mislocalize concepts to backgrounds.
   In contrast, our concept-level explanations are not only visually well-disentangled, but also correctly localized.
   This indicates that our model understands the semantic meanings of fine-grained concepts, and can visually distinguish them.
   }
   \label{fig:exp_per_image}
\end{figure}

\begin{figure*}[t]
  \centering
   \includegraphics[width=1.0\linewidth]{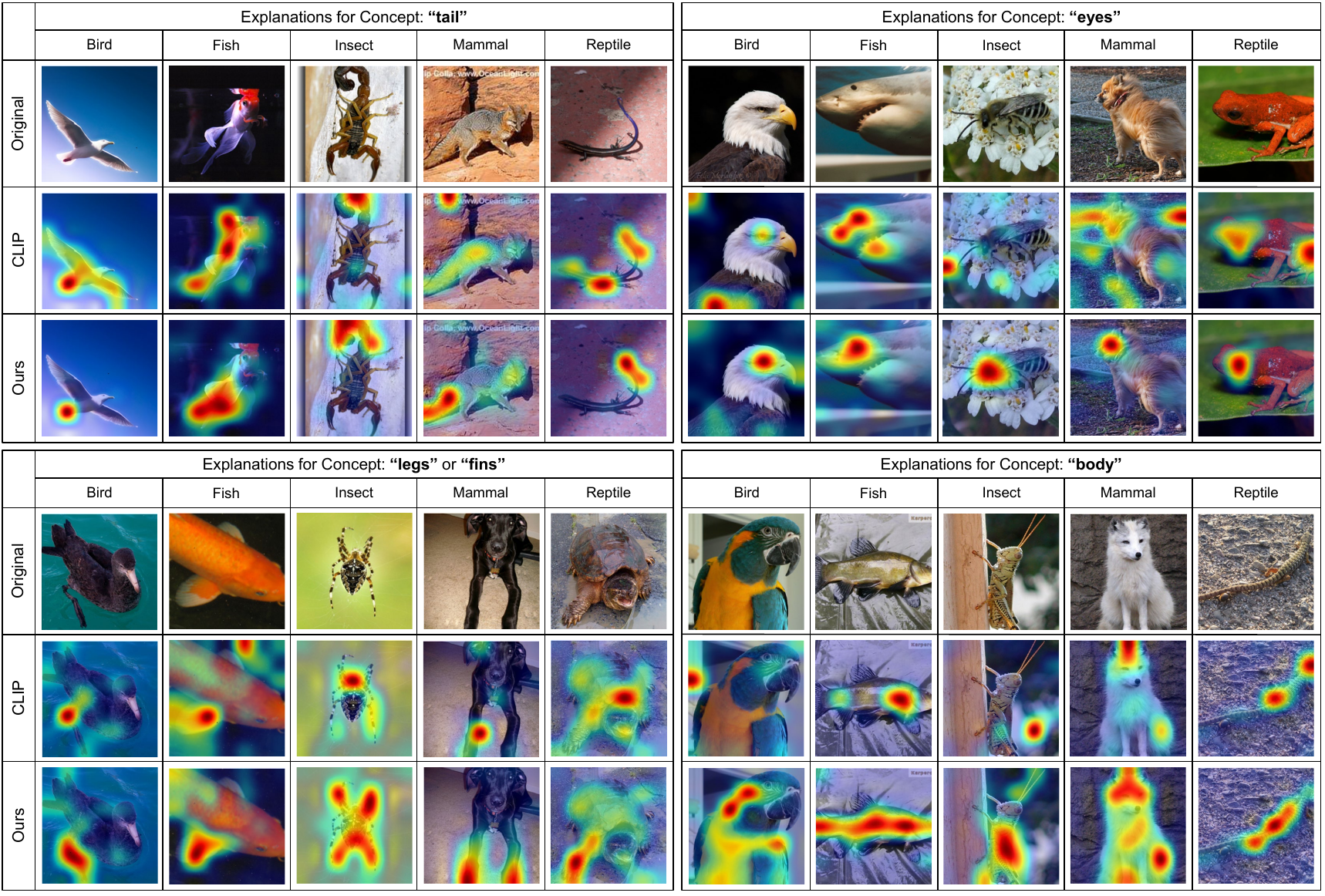}
   \caption{
   Localizability comparison per concept using different animal species groups from the {\it ImageNet}~\cite{deng2009imagenet} dataset.
   The CLIP~\cite{radford2021learning} model's explanations {\it w.r.t.} the same concept might vary across different species, they usually entangle with other concepts or even spurious factors.
   In contrast, our model consistently localizes the region {\it w.r.t.} the semantic meaning of the concept.
   Note that we explicitly select images from five diverse animal species groups, {\it i.e.}, Bird, Fish, Insect, Mammal, and Reptile.
   The results indicate that our model can generalize the learned concepts across categories even though their appearance drastically changed.
   This figure is best viewed in color.
   }
   \label{fig:exp_per_concept}
\end{figure*}

\subsection{Evaluation Metrics}

{\bf Concept-level Explanation Disentanglability.}
The disentanglability of concept-level explanations is evaluated through the $\mathcal{R}_\mathrm{disen}$ regularization term that is also applied in our model's training phase.
While this term was optimized on training data, the disentanglability of concept-level explanations on the hold-out test data remains essential for evaluating the success of our goal.
For better comparison, its values are normalized to [0, 1].
A higher metric score is expected for concept-level explanations that exhibit better disentanglement.

{\bf Concept-level Explanation Localizability.}
Due to the absence of region annotations for concepts, we use the explanation fidelity~\cite{Petsiuk2018rise} {\it w.r.t.} the concepts to measure their localizability.
This metric quantifies the increase in predictive score probability with the sequential inclusion of pixels, sorted in descending order of their importance as derived from the explanation~\cite{10.5555/3491440.3491857}.
Different from the conventional category-level version, our metric measures the fidelity at the concept level using the predictive score of concepts.
We calculate the area under the insertion curve (iAUC), capturing the cumulative probability increase as features from the original input are incrementally inserted into a reference input.
We employ a blurred canvas as the reference input to minimize the introduction of distributional shifts.
To compare across models, we follow~\cite{schulzrestricting} to normalize iAUC scores to [0, 1].
A higher iAUC score is expected for an explanation that is faithful with respect to the concept.

{\bf Prediction Accuracy.}
The typical practice for VLMs to make predictions is to directly rank the similarity between the image and all category name embeddings, and select the top-1 category as prediction~\cite{radford2021learning}.
In contrast, to better evaluate the model's understanding of fine-grained concepts, we propose using the similarity between the image and all concept embeddings to make predictions.
Specifically, as described in Sec.~\ref{sec:prompt}, each category consists of a set of $K$ concepts.
The category similarity will be determined by the average embedding similarity between the image and all concepts within that category:
\begin{equation}
s(I, \texttt{[CATEGORY]}) = \frac{1}{K} \sum_{k=1}^{K} \mathrm{similarity} (f(I, C_k))
\label{eq:concept_pred}
\end{equation}
This has been proved by recent works~\cite{menon2022visual, qin2022medical} that can improve the zero-shot prediction performance of VLMs.
In this paper, we use it as a natural metric.

\subsection{Comparison with State-of-the-Art VLMs}
\noindent {\bf Takeaway 1: DEAL can disentangle and localize fine-grained concepts.}
As aforementioned, the existing learning objectives of VLMs are limited to the category (or object) level alignment, while our method achieves finer alignment at the concept level.
Furthermore, existing works typically rely on expensive annotations of objects, while our method leverages self-supervised learning thereby eliminating the need for human annotations.
In this section, we compare our method with representative VLMs ({\it i.e.}, CLIP~\cite{radford2021learning}, FLAVA~\cite{singh2022flava}, and DeCLIP~\cite{li2021supervision}) and state-of-the-art VLMs using for category-level alignment ({\it i.e.}, PyramidCLIP~\cite{gao2022pyramidclip} and CLIPpy~\cite{ranasinghe2023perceptual}).

{\bf Evaluation on disentanglability.}
Quantitatively, our model significantly outperforms VLMs and their fine-tuned versions in terms of disentanglablity.
As shown in Tab.~\ref{tab:experiments}, on the average of five benchmark datasets, our method outperforms the second-best results by 8.8\%.
Qualitatively, in Fig.~\ref{fig:exp_per_image}, we visualize five different concept-level explanations for each image from the {\it ImageNet}~\cite{deng2009imagenet} dataset.
As shown, the concept-level explanations of the CLIP model are entangled, exhibiting a high degree of similarity between them.
In contrast, the concept-level explanations of our model are visually disentangled.
Furthermore, our concept-level explanations accurately segment the contour of the corresponding concepts, even though most concepts lack clear borders with adjacent concepts.
Our results reveal that the global alignment learning objectives of existing VLMs entangle the model's understanding of fine-grained concepts.

{\bf Evaluation on localizability.}
Quantitatively, as shown in Tab.~\ref{tab:experiments}, our model outperforms VLMs and their fine-tuned versions in terms of localizability.
On the average of five benchmark datasets, our method outperforms the second-best results by 10.9\%.
Qualitatively, in Fig.~\ref{fig:exp_per_concept}, we visualize the explanations of five different images for each concept.
The images are from five diverse animal species groups in the {\it ImageNet}~\cite{deng2009imagenet} dataset.
As shown, the CLIP model usually associates the concepts with the entire object, sometimes even highlighting the spurious correlations ({\it e.g.}, watermarks).
On the contrary, our model can differentiate fine-grained concepts and accurately highlight regions that align with the concepts' semantic meanings.
The results demonstrate the superior capability of our model in understanding fine-grained concepts, even though the presentation of these concepts significantly differs among species groups.

{\bf Evaluation on prediction accuracy.}
Our model, as shown in Tab.~\ref{tab:experiments}, outperforms state-of-the-art VLMs in terms of prediction accuracy.
On average of five benchmark datasets, our method outperforms the second-best results by 13.1\%.
This finding is surprising since interpretability methods usually come with a compromise on the benchmark performance~\cite{rudin2019stop, chen2019looks}.
The result indicates that, through improving concept-level explanations, our method alleviates the model's reliance on spurious correlations~\cite{arjovsky2019invariant}, {\it e.g.}, watermark highlights in Fig.~\ref{fig:exp_per_concept}, thereby benefiting its prediction performance on unseen data.

\begin{table}[t]
\centering
\caption{
Comparison with typical fine-tuning on CLIP~\cite{radford2021learning} model.
The abbreviations ``zs'', ``ft-full'', and ``ft-vision'' correspond to the zero-shot, full model fine-tuning, and vision-encoder-only fine-tuning.
Our results demonstrate that adapting pretrained VLMs to downstream tasks without constraints might deteriorate the model's understanding of concepts and yield compromised performance.
In contrast, our method achieves superior performance through the disentanglement and localization of fine-grained concepts.
$^\dagger$ Reported in~\cite{radford2021learning}.
See Supplementary for results of other datasets.
}
\resizebox{1.0\columnwidth}{!}{%
\begin{tabular}{lcccccc}
\toprule[1pt]
\multicolumn{1}{c}{\multirow{2}{*}{Models}} & \multicolumn{3}{c}{ImageNet}                                                        & \multicolumn{3}{c}{CUB}                                                  \\ \cmidrule{2-7} 
\multicolumn{1}{c}{}                        & Disentanglability $\uparrow$ \; & Localizability $\uparrow$ & \; Acc. \;                     & Disentanglability $\uparrow$ \; & Localizability $\uparrow$ \; & Acc.          \\ \midrule[0.75pt]
RN50-zs                                     & {\ul 0.677}                  & 0.608                     & \multicolumn{1}{l}{$^\dagger$59.6} & {\ul 0.679}                  & 0.613                     & 48.7          \\
RN50-ft-full                                & 0.452                        & 0.641                     & {\ul 68.8}                     & 0.608                        & {\ul 0.647}               & {\ul 55.0}          \\
RN50-ft-vision                              & 0.572                        & {\ul 0.642}               & 61.8                     & 0.641                        & 0.615                     & 54.7          \\
RN50-ours                                   & \textbf{0.690}               & \textbf{0.661}            & \textbf{70.0}            & \textbf{0.725}               & \textbf{0.651}            & \textbf{57.4} \\ \toprule[0.5pt]
ViT-B/32-zs                                 & {\ul 0.361}                  & {\ul 0.633}               & \multicolumn{1}{l}{$^\dagger$63.2} & {\ul 0.596}                  & {\ul 0.638}               & 52.6          \\
ViT-B/32-ft-full                             & 0.323                        & 0.618                     & {\ul 68.9}                     & 0.579                        & 0.602                     & {\ul 67.5}          \\
ViT-B/32-ft-vision                          & 0.207                        & 0.617                     & 66.4                     & 0.549                        & 0.624                     & 62.1          \\
ViT-B/32-ours                               & \textbf{0.397}               & \textbf{0.673}            & \textbf{70.8}            & \textbf{0.608}               & \textbf{0.718}            & \textbf{69.6} \\ \bottomrule[1pt]
\end{tabular}
}
\label{tab:fine-tune}
\end{table}

\subsection{Comparison with Fine-tuned Models}
\noindent {\bf Takeaway 2: The disentanglability of DEAL can boost prediction accuracy.}
To better evaluate the effectiveness of the proposed constraints, we compare our method with zero-shot and representative fine-tuning methods ({\it i.e.}, full model fine-tuning, and vision-encoder-only fine-tuning of CLIP) on the five benchmark datasets.
We use Grad-CAM~\cite{selvaraju2017grad} for ResNet-50 and Chefer et al.~\cite{Chefer_2021_ICCV} for ViT-B/32 to generate explanations {\it w.r.t.} concepts.
As shown in Tab.~\ref{tab:fine-tune}, our method outperforms the second-best results by 6.8\%, 5.5\%, and 2.1\% on average of five datasets and two backbones in terms of disentanglability, localizability, and prediction accuracy. See Supplementary for the full table and details.

\begin{table}[t]
\centering
\caption{
Evaluation using ground truth object part annotations.
For all models, the mean Intersection over Union (mIoU) is calculated between the explanations of part and their corresponding ground truth segmentation masks.
The results are on the average of 10 different parts for the {\it CUB-part} and 40 parts for the {\it PartImageNet} datasets.
}
\resizebox{0.65\columnwidth}{!}{%
\begin{tabular}{lccccccc}
\toprule[0.75pt]
\multicolumn{1}{c}{\multirow{2}{*}{Datasets}} & \multicolumn{3}{c}{ViT-B/32}                         &  & \multicolumn{3}{c}{RN50}        \\ \cmidrule{2-4} \cmidrule{6-8} 
\multicolumn{1}{c}{}                         & CLIP                      & CLIP-ft & Ours           &  & CLIP & CLIP-ft & Ours           \\ \toprule[0.5pt]
CUB-Part~\cite{saha2022improving}            & \multicolumn{1}{l}{10.49} & 10.24   & \textbf{15.69} &  & 6.31 & 6.52    & \textbf{12.04} \\
PartImageNet~\cite{he2022partimagenet}       & 8.79                      & 7.25    & \textbf{12.73} &  & 4.92 & 5.19    & \textbf{10.04} \\ \bottomrule[0.75pt]
\end{tabular}
}
\label{tab:parts}
\end{table}

\subsection{Evaluation using Ground Truth Object Part Annotations}
\noindent {\bf Takeaway 3: DEAL can accurately localize fine-grained object parts.}
Object parts represent a specific subset of fine-grained concepts that can evaluate the model's localization capability.
We evaluate our model on the {\it CUB-Part}~\cite{saha2022improving} and {\it PartImageNet}~\cite{he2022partimagenet} datasets that provide ground truth masks of parts, such as ``beak'' and ``head'', for 299 and 24,000 images.
As shown in Tab.~\ref{tab:parts}, our model significantly improves the localization of explanations for these concepts by 5.20\% and 5.52\% for two different vision backbones on {\it CUB-Part}, and 3.94\% and 4.85\% on {\it PartImageNet} in terms of mIoU score.

\subsection{Ablation Study}
\noindent In this section, we perform ablation studies to investigate the key components of our method proposed in Eq.~(\ref{eq:disent}) and Eq.~(\ref{eq:consis}).
The empirical results are shown in Tab.~\ref{tab:ablation}.
Furthermore, as shown in Tab.~\ref{tab:scalability}, we compare the scalability between our method and the baseline fine-tuning method.

{\bf Ablation on disentanglement constraint.}
The ``DEAL w/o disen.'' represents a variant of our method which only optimizes the consistency between the aggregation of concept-level explanations and the corresponding category-level explanation.
Without constraints on the disentanglement, the concept-level explanations tend to be similar to easily satisfy the consistency constraint.
This leads to entangled concept-level explanations, which also dramatically decrease the prediction performance by 10.4\%.

{\bf Ablation on localization constraint.}
The ``DEAL w/o local.'' represents a variant of our method which only optimizes for the disentanglement of explanations.
As we discussed in Sec.~\ref{sec:methods}, recklessly disentangling concept-level explanation might easily fall into a local minimum.
As shown in Tab~\ref{tab:ablation}, although achieving high disentanglement, such explanations are incorrectly localized (low localizability) and significantly compromise the prediction performance by 8.1\%.

{\bf Scalability comparison.}
We compare the scalability of our method with standard training.
Training the full CLIP~\cite{radford2021learning} with ViT-B/32 encoder on the {\it ImageNet}~\cite{deng2009imagenet} dataset for one epoch using our method requires 8 hours and 6 minutes of wall clock time, compared to 4 hours and 6 minutes for standard training.
However, this additional time investment is justified by the benefits.
As shown in Tab.~\ref{tab:scalability}, our method significantly improves the model's concept-level explanations in terms of disentanglability and localizability by 22.9\% and 8.9\%.
Without extra parameters, our method improves the accuracy by 1.9\%.

\begin{table}[t]
\centering
\caption{
Ablation study on {\it ImageNet}~\cite{deng2009imagenet} dataset.
In this study, we use the ViT-B/32 vision backbone.
The variant without disentanglement produces entangled explanations thereby drastically deteriorating prediction accuracy.
The variant without localization recklessly optimizes for disentanglement, which falls into a local minimum with compromised localizability and prediction accuracy.
}
\resizebox{0.65\columnwidth}{!}{%
\begin{tabular}{lccc}
\toprule[1pt]
Models          & Disentanglability $\uparrow$\; & Localizability $\uparrow$\; & Acc. (\%)   \\ \toprule[0.75pt]
CLIP~\cite{radford2021learning}            & 0.361{\footnotesize $\pm$0.00}               & 0.633{\footnotesize $\pm$0.00}            & 63.2{\footnotesize $\pm$0.0}          \\
CLIP-ft-full    & 0.323{\footnotesize $\pm$0.01}               & 0.618{\footnotesize $\pm$0.01}            & {\ul 68.9}{\footnotesize $\pm$0.3}    \\ \midrule[0.5pt]
DEAL w/o disen. & 0.354{\footnotesize $\pm$0.01}               & {\ul 0.657}{\footnotesize $\pm$0.02}      & 60.4{\footnotesize $\pm$0.8}          \\
DEAL w/o local. & \textbf{0.702}{\footnotesize $\pm$0.03}      & 0.623{\footnotesize $\pm$0.01}            & 62.7{\footnotesize $\pm$0.6}          \\
DEAL (Ours)     & {\ul 0.397}{\footnotesize $\pm$0.01}         & \textbf{0.673}{\footnotesize $\pm$0.01}   & \textbf{70.8}{\footnotesize $\pm$0.4} \\ \bottomrule[1pt]
\end{tabular}
}
\label{tab:ablation}
\end{table}

\begin{table}[t]
\centering
\caption{
Scalability comparison for training one epoch on {\it ImageNet}~\cite{deng2009imagenet} dataset.
In this comparison, we use the ViT-B/32 vision backbone.
Note that our training time varies depending on the number of concepts (we use five in this example).
The time investment is justified by the benefits, our method significantly outperforms standard training in terms of explanation quality and prediction performance.
}
\resizebox{0.7\columnwidth}{!}{%
\begin{tabular}{lccccc}
\toprule[1pt]
Models       & \#Param.  & Time  & Disentangle $\uparrow$\; & Localize $\uparrow$\; & Acc (\%) \\ \toprule[0.75pt]
CLIP~\cite{radford2021learning}         & 151M      & -     & 0.361                  & 0.633               & 63.2                \\
CLIP-ft-full & 151M      & 4h06m & 0.323                  & 0.618               & 68.9                \\
DEAL (Ours)  & 151M      & 8h06m & \textbf{0.397}         & \textbf{0.673}      & \textbf{70.8}       \\ \bottomrule[1pt]
\end{tabular}
}
\label{tab:scalability}
\end{table}

\section{Conclusion}
\label{sec:conclusion}
\noindent In this paper, we present comprehensive studies to show that the training objective of existing VLMs ignores the distinction between fine-grained concepts.
Their explanations with respect to these concepts are entangled and mislocalized.
To address this problem, we propose to Disentangle and Localize (DEAL) the concept-level explanations for VLMs.
Our method fully utilizes the discrepancy and consistency between concept- and category-level explanations to provide supervisory signals for concept learning without human annotations.
Extensive experiments demonstrate the superiority of our method in terms of the disentanglability and localizability of concept-level explanations.

\section*{Acknowledgment} 
\noindent This work is supported by the National Science Foundation through the Faculty Early Career Development Program (NSF CAREER) Award NSF-IIS-2340074 and the Department of Defense under the Defense Established Program to Stimulate Competitive Research (DoD DEPSCoR) Award.


%
%
\bibliographystyle{splncs04}
\bibliography{main}

\title{Supplementary Material for DEAL: Disentangle and Localize Concept-level Explanations for VLMs} 

\titlerunning{Supplementary Material for DEAL}

\author{Tang Li \orcidlink{0000-0002-3134-4151} \and
Mengmeng Ma \orcidlink{0000-0002-2804-2718} \and
Xi Peng \orcidlink{0000-0002-7772-001X}}

\authorrunning{T. Li et al.}

\institute{Deep-REAL Lab: \url{https://deep-real.github.io/} \\
Department of Computer \& Information Science, University of Delaware \\
\email{\{tangli,mengma,xipeng\}@udel.edu}}

\maketitle



\section{Additional Results}

\subsection{Comparison with Fine-tuned Models}
To better evaluate the effectiveness of the proposed constraints, we compare our method with zero-shot and representative fine-tuning methods ({\it i.e.}, full model fine-tuning, and vision-encoder-only fine-tuning of CLIP) on the five benchmark datasets.
We use Grad-CAM~\cite{selvaraju2017grad} for ResNet-50 and~\cite{Chefer_2021_ICCV} implementation for ViT-B/32 to generate explanations {\it w.r.t.} concepts.
Here we provide our results of all datasets in Tab.~\ref{tab:full_table} as the full table of Tab.2 in the main paper.
As shown, our model outperforms standard fine-tuning models in terms of explanation and prediction performance.

\begin{table}[t]
    \centering
    \caption{Comparison of the explanation and prediction performance on {\it ImageNet}~\cite{deng2009imagenet}, {\it CUB}~\cite{wah2011caltech}, {\it Food-101}~\cite{bossard14}, {\it Oxford-Pets}~\cite{parkhi12a}, and {\it EuroSA}T~\cite{helber2019eurosat} datasets.
    The abbreviations ``zs'', ``ft-full'', and ``ft-vision'' correspond to the zero-shot, full model fine-tuning, and vision-encoder-only fine-tuning performance of the CLIP~\cite{radford2021learning} model.
    The models are tested on hold-out testing data.
    Our results are on the average of three trials of experiments.
    We highlight the {\bf best results} and the \underline{second best} results.
    Note that the zero-shot prediction accuracy of ViT-B/32 is from~\cite{menon2022visual}.
    On average, our method exhibits superior concept-level explanation in terms of disentanglability and localizability.
    The improved explainability also benefits the model's prediction accuracy.
    $^\dagger$ Reported in~\cite{radford2021learning}.
    }
\resizebox{1.0\columnwidth}{!}{%
\begin{tabular}{clcccccc}
\toprule[1.2pt]
\multirow{2}{*}{Metrics}                                                                                            & \multicolumn{1}{c}{\multirow{2}{*}{Models}} & \multicolumn{6}{c}{Datasets}                                                                        \\ \cline{3-8} 
                                                                                                                    & \multicolumn{1}{c}{}                         & ImageNet       & CUB            & Food-101       & Oxford-Pets    & EuroSAT        & Avg.           \\ \toprule[1pt]
\multirow{8}{*}{\begin{tabular}[c]{@{}c@{}}Concept-level Explanation\\ Disentanglability $\uparrow$  \end{tabular}} & RN50-zs                                      & {\ul 0.677}    & {\ul 0.679}    & {\ul 0.327}    & 0.176          & {\ul 0.705}    & {\ul 0.513}    \\
                                                                                                                    & RN50-ft-full                                 & 0.452          & 0.608          & 0.196          & {\ul 0.201}    & 0.544          & 0.400          \\
                                                                                                                    & RN50-ft-vision                               & 0.572          & 0.641          & 0.224          & 0.186          & 0.520          & 0.429          \\
                                                                                                                    & RN50-ours                                    & \textbf{0.690} & \textbf{0.725} & \textbf{0.366} & \textbf{0.204} & \textbf{0.713} & \textbf{0.540} \\ \cline{2-8} 
                                                                                                                    & ViT-B/32-zs                                  & {\ul 0.361}    & {\ul 0.596}    & {\ul 0.487}    & 0.363          & \textbf{0.192} & {\ul 0.400}    \\
                                                                                                                    & ViT-B/32-ft-full                             & 0.323          & 0.579          & 0.430          & 0.428          & 0.190          & 0.390          \\
                                                                                                                    & ViT-B/32-ft-vision                           & 0.207          & 0.549          & 0.457          & {\ul 0.430}    & 0.175          & 0.364          \\
                                                                                                                    & ViT-B/32-ours                                & \textbf{0.397} & \textbf{0.608} & \textbf{0.501} & \textbf{0.475} & \textbf{0.192} & \textbf{0.435} \\ \toprule[1pt]
\multirow{8}{*}{\begin{tabular}[c]{@{}c@{}}Concept-level Explanation\\ Localizability $\uparrow$\end{tabular}}    & RN50-zs                                      & 0.608          & 0.613          & \textbf{0.550} & 0.668          & 0.344          & 0.557          \\
                                                                                                                    & RN50-ft-full                                 & 0.641          & {\ul 0.647}    & 0.523          & 0.690          & {\ul 0.535}    & {\ul 0.607}    \\
                                                                                                                    & RN50-ft-vision                               & {\ul 0.642}    & 0.615          & 0.515          & {\ul 0.694}    & 0.522          & 0.598          \\
                                                                                                                    & RN50-ours                                    & \textbf{0.661} & \textbf{0.651} & {\ul 0.547}    & \textbf{0.705} & \textbf{0.559} & \textbf{0.625} \\ \cline{2-8} 
                                                                                                                    & ViT-B/32-zs                                  & {\ul 0.633}    & {\ul 0.638}    & 0.511          & 0.762          & {\ul 0.423}    & 0.593          \\
                                                                                                                    & ViT-B/32-ft-full                             & 0.618          & 0.602          & {\ul 0.612}    & {\ul 0.807}    & 0.422          & {\ul 0.612}    \\
                                                                                                                    & ViT-B/32-ft-vision                           & 0.617          & 0.624          & 0.589          & 0.737          & 0.405          & 0.594          \\
                                                                                                                    & ViT-B/32-ours                                & \textbf{0.673} & \textbf{0.718} & \textbf{0.660} & \textbf{0.809} & \textbf{0.444} & \textbf{0.661} \\ \toprule[1pt]
\multirow{8}{*}{\begin{tabular}[c]{@{}c@{}}Prediction Accuracy (\%)                           \end{tabular}}        & RN50-zs                                      & $^\dagger$59.6           & 48.7           & $^\dagger$81.1           & $^\dagger$85.4           & $^\dagger$41.1           & 63.2           \\
                                                                                                                    & RN50-ft-full                                 & {\ul 68.8}     & {\ul 55.0}     & {\ul 83.3}     & {\ul 89.4}     & 77.9           & {\ul 74.9}     \\
                                                                                                                    & RN50-ft-vision                               & 61.8           & 54.7           & 79.2           & 79.1           & {\ul 82.5}     & 71.5           \\
                                                                                                                    & RN50-ours                                    & \textbf{70.0}  & \textbf{57.4}  & \textbf{83.4}  & \textbf{89.5}  & \textbf{84.8}  & \textbf{77.0}  \\ \cline{2-8} 
                                                                                                                    & ViT-B/32-zs                                  & $^\dagger$63.2           & 52.6           & $^\dagger$84.4           & $^\dagger$87.0           & $^\dagger$49.4           & 67.3           \\
                                                                                                                    & ViT-B/32-ft-full                             & {\ul 68.9}     & {\ul 67.5}     & {\ul 86.5}     & {\ul 89.1}     & {\ul 76.4}     & {\ul 77.7}     \\
                                                                                                                    & ViT-B/32-ft-vision                           & 66.4           & 62.1           & 83.1           & 86.6           & 75.6           & 74.8           \\
                                                                                                                    & ViT-B/32-ours                                & \textbf{70.8}  & \textbf{69.6}  & \textbf{86.9}  & {\ul 89.3}     & \textbf{77.4}  & \textbf{78.8}  \\ \bottomrule[1.2pt]
\multicolumn{1}{l}{}                                                                                                &                                              &                &                &                &                &                &               
\end{tabular}
}
\label{tab:full_table}
\end{table}

\subsection{Concept-level Explanations per Image}
\begin{figure*}[t]
  \centering
   \includegraphics[width=1.0\linewidth]{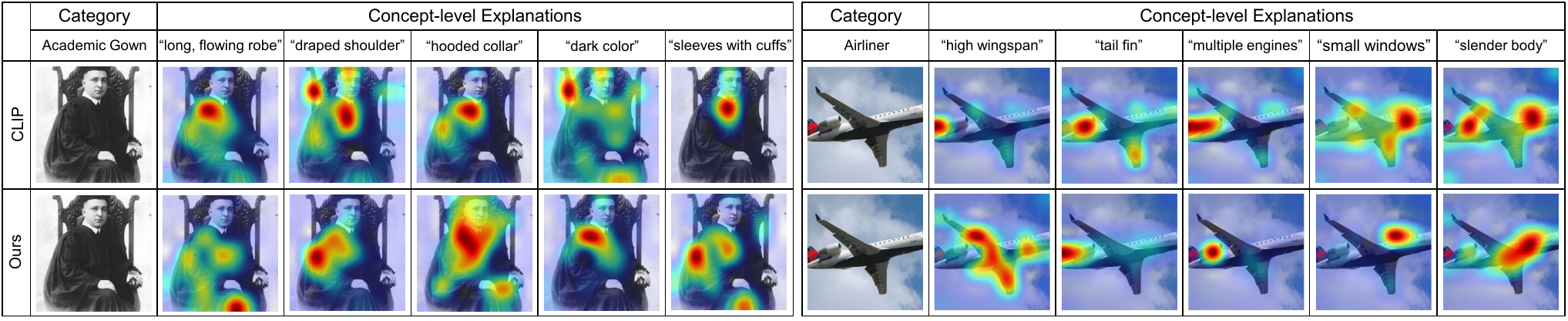}

   \caption{
   Disentanglability comparison using images from the {\it ImageNet}~\cite{deng2009imagenet} dataset.
   This figure is best viewed in color.
   }
   \label{fig:exp_per_image_sup}
\end{figure*}

In the main paper, we show the disentanglability of our concept-level explanations for different concepts of the same image.
In addition to animal images, we visualize the concept-level explanations per image from other non-animal categories.
As shown in Fig.~\ref{fig:exp_per_image_sup}, the CLIP~\cite{radford2021learning} model's explanations {\it w.r.t.} different concepts are entangled.
They might highlight the same region for visually distinct concepts, or even mislocalize concepts to backgrounds.
In contrast, our concept-level explanations are not only visually well-disentangled, but also correctly localized.
This indicates that our model understands the semantic meanings of fine-grained concepts, and can visually distinguish them.
For example, on the right side of Fig.~\ref{fig:exp_per_image_sup}, the CLIP's explanation of ``high wingspan'' is focused on the tail fin part of the airliner, in contrast, our explanation clearly highlights the contour of the wings of the airliner.

\subsection{Concept-level Explanations per Concept}
\begin{figure*}[t]
  \centering
   \includegraphics[width=1.0\linewidth]{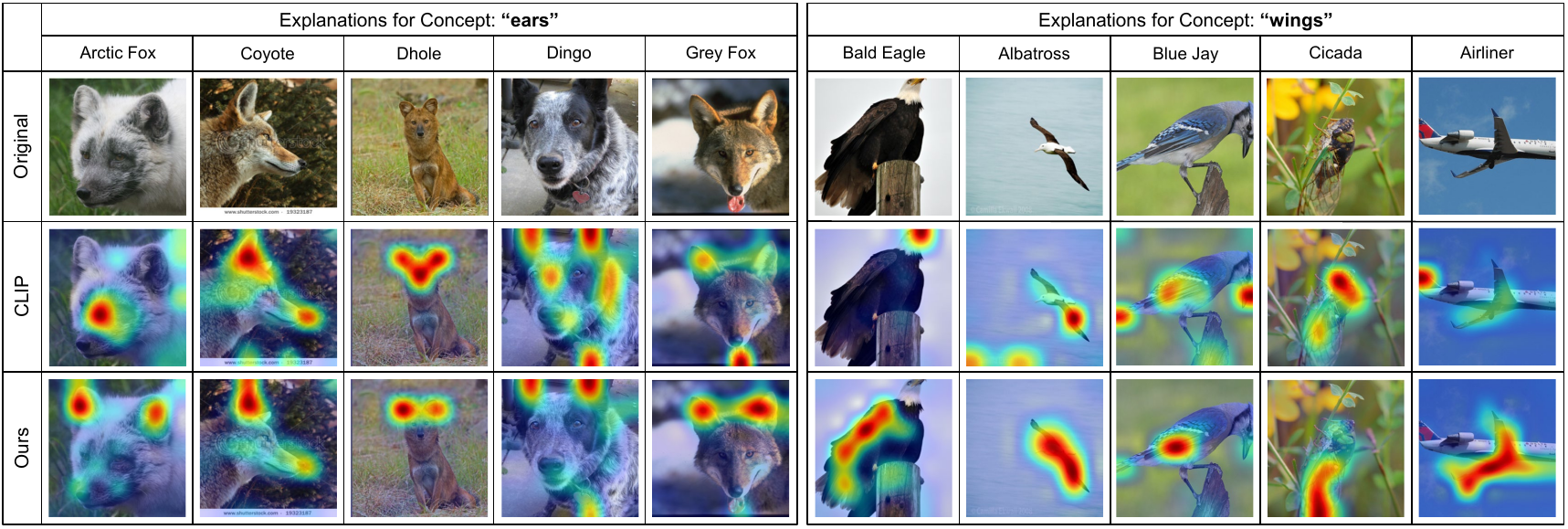}

   \caption{
   Localizability comparison per concept using images from the {\it ImageNet}~\cite{deng2009imagenet} dataset.
   This figure is best viewed in color.
   }
   \label{fig:exp_per_concept_sup}
\end{figure*}

In the main paper, we show the localizability of our concept-level explanations for different concepts that are shared across species groups.
In addition to those commonly shared concepts, there are also concepts that are only shared between some of the species or other non-animal categories.
As shown in Fig~\ref{fig:exp_per_concept_sup}, we show the concept ``ears'' that are only visually distinctive for mammals, and ``wings'' that are shared across birds, insects, and airliners.
The CLIP~\cite{radford2021learning} model's explanations {\it w.r.t.} the same concept might vary across different categories, they usually entangle with other concepts or even spurious factors.
In contrast, our model consistently localizes the region {\it w.r.t.} the semantic meaning of the concept.
For example, although the appearance of the concept ``wings'' is drastically changed between birds, insects, and airliners, our concept-level explanations can still accurately localize the wing region.
This indicates that our model understands the true meaning of the concept even though its presentation drastically changed across categories.

\begin{figure*}[t]
  \centering
   \includegraphics[width=1.0\linewidth]{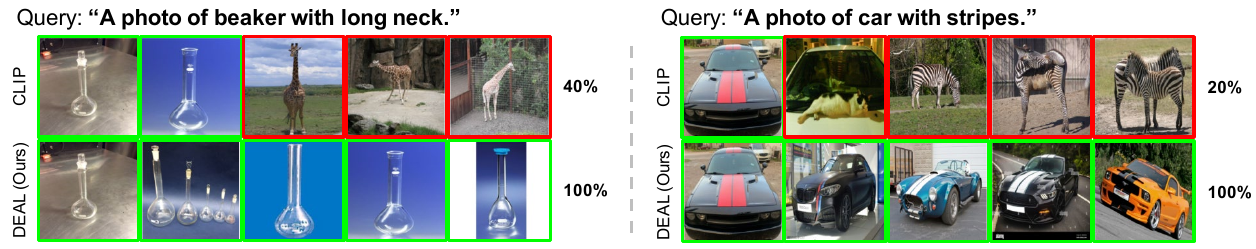}
\vspace{-15pt}
\caption{
   Top-5 concept-based text-to-image retrieval results comparison using ViT-B/32 on {\it MSCOCO}~\cite{lin2014microsoft} dataset.
   ({\it Left}) the CLIP model entangles the concept ``long neck'' tightly with giraffe.
   ({\it Right}) the CLIP model entangles the concept ``stripe'' with zebra.
   }
   \label{fig:retrieval}
\vspace{-10pt}
\end{figure*}

\subsection{Evaluation on Retrieval Tasks}
In order to better evaluate the superiority of our learned representations on downstream tasks, we introduce a novel and more challenging retrieval task, {\it concept-based text-to-image retrieval}.
The model is tasked with retrieving images according to descriptions that detail an object with fine-grained attributes.
Specifically, five new images characterized by specific attributes are incorporated into the existing 5,000-image {\it MSCOCO}~\cite{lin2014microsoft} validation set, ensuring that objects with such attributes have not been represented in the original dataset.
As shown in Fig~\ref{fig:retrieval}, for both categories, our model achieves 100\% recall, while CLIP~\cite{radford2021learning} only obtains 40\% and 20\% respectively.
Furthermore, the visualizations indicate that the CLIP model entangles certain attributes with specific categories.

\section{Experimental Details}

We provide additional experimental details in Tab~\ref{tab:hyper}.
For all datasets, our hyperparameters are consistent.
We use the implementations of CLIP~\cite{radford2021learning} for different vision backbones.
Due to the computational cost of training CLIP from scratch, we focused on finetuning experiments on ViT-B/32 and ResNet-50 backbones.
Note that the random distribution shows the range of our parameter searching.
\begin{table}[t]
\centering
\caption{
Hyperparameters, default values, and their distributions.
}
\resizebox{0.7\columnwidth}{!}{%
\begin{tabular}{llll}
\toprule[1pt]
Vision Backbone           & Parameter     & Default Values             & Random Distribution           \\ \toprule[0.75pt]
\multirow{8}{*}{RN50}     & learning rate & 5e-5                       & $10^\mathrm{Uniform(-6, -5)}$ \\
                          & batch size    & 256                        & $2^\mathrm{Uniform(5, 9)}$    \\
                          & lambda        & 0.1                        & $10^\mathrm{Uniform(-3, 1)}$  \\
                          & gamma         & 0.01                       & $10^\mathrm{Uniform(-3, 1)}$  \\
                          & optimizer     & Adam~\cite{kingma2014adam} & -                             \\
                          & weight decay  & 0.001                      & $10^\mathrm{Uniform(-4, -3)}$ \\
                          & betas         & (0.9, 0.98)                & -                             \\
                          & eps           & 1e-6                       & $10^\mathrm{Uniform(-8, -6)}$ \\ \toprule[0.75pt]
\multirow{8}{*}{ViT-B/32} & learning rate & 5e-5                       & $10^\mathrm{Uniform(-6, -5)}$ \\
                          & batch size    & 256                        & $2^\mathrm{Uniform(5, 9)}$    \\
                          & lambda        & 0.05                       & $10^\mathrm{Uniform(-3, 1)}$  \\
                          & gamma         & 0.01                       & $10^\mathrm{Uniform(-3, 1)}$  \\
                          & optimizer     & Adam~\cite{kingma2014adam} & -                             \\
                          & weight decay  & 0.001                      & $10^\mathrm{Uniform(-4, -3)}$ \\
                          & betas         & (0.9, 0.98)                & -                             \\
                          & eps           & 1e-6                       & $10^\mathrm{Uniform(-8, -6)}$ \\ \bottomrule[1pt]
\end{tabular}
}
\label{tab:hyper}
\end{table}

\section{Related Works}

{\bf Explanation-guided learning.}
Recent efforts have been made to incorporate prediction explanations into model training to improve predictive performance.
\cite{rieger2020interpretations, stammer2021right} match model-generated prediction explanations with human-annotated ground truth explanations based on domain expertise, directing the model’s focus toward foreground objects rather than the background pixels.
However, these annotations are limited to the object level, acquiring more detailed annotations at the fine-grained concept level is prohibitively labor-intensive or even impractical~\cite{wang2020self, roscher2020explainable}.
There are recent attempts to leverage Self-supervised Learning to provide supervisory signals to constrain the explanations.
\cite{guo2019visual, pillai2022consistent, li2023data} align explanations between image transformations or data distributions to improve image classification and weakly-supervised segmentation.
\cite{chen2019robust, han2021explanation, cugu2022attention} enhancing model robustness by ensuring consistency in explanations across perturbed and original samples.
However, such feature importance-based explanations are usually ambiguous and require machine learning expertise to interpret.
In contrast to existing methods, the proposed method leverages the disentanglement between concept-level explanations to provide supervisory signals without human annotations.
Furthermore, our method provides user-understandable concepts and valid visual evidence for them to explain model predictions.


%
%

\end{document}